\documentclass[letterpaper, 10 pt, conference]{ieeeconf}  

\IEEEoverridecommandlockouts                              

\usepackage{graphicx} 
\usepackage{epsfig} 
\usepackage{amsmath} 
\usepackage{algorithm}
\usepackage{algpseudocode}
\usepackage{caption}
\usepackage{subcaption}
\usepackage{amssymb}
\usepackage{booktabs}
\usepackage{multirow}
\usepackage{balance}
\usepackage{array}
\usepackage{cellspace}
\usepackage{threeparttable}

\graphicspath{{images/}}

\title{\LARGE \bf
EHC-MM: Embodied Holistic Control for Mobile Manipulation  
}

\author{Jiawen Wang$^{1,2}$, Yixiang Jin$^{1}$, Jun Shi$^{1}$, Yong A$^{1}$, Dingzhe Li$^{1}$, Fuchun Sun$^{3}$, Dingsheng Luo$^{2\dag}$, Bin Fang$^{4\dag}$
\thanks{$^{1}$The authors are with Beijing Samsung Telecom R\&D Center.}
\thanks{$^{2}$J. Wang and D. Luo are with the National Key Laboratory of General Artificial Intelligence \& School of Intelligence Science and Technology, Peking University \& PKU-WUHAN Institute for Artificial Intelligence \& Institute for Artificial Intelligence, Peking university (e-mail: \textit{chiawenw@stu.pku.edu.cn}, \textit{dsluo@pku.edu.cn}).}
\thanks{$^{3}$The author is with Tsinghua University.}%
\thanks{$^{4}$B. Fang is with Beijing University of Posts and Telecommunications (e-mail: \textit{fangbin1120@bupt.edu.cn}).}
\thanks{$^{\dag}$The authors are co-corresponding authors.}
}

\begin{document}

\maketitle
\thispagestyle{empty}
\pagestyle{empty}

\begin{abstract}
Mobile manipulation typically entails the base for mobility, the arm for accurate manipulation, and the camera for perception. The principle of \textit{Distant Mobility, Close Grasping}(DMCG) is essential for holistic control. We propose Embodied Holistic Control for Mobile Manipulation(EHC-MM) with the embodied function of $\textbf{sig}(\omega)$: By formulating the DMCG principle as a Quadratic Programming (QP) problem, $\textbf{sig}(\omega)$ dynamically balances the robot’s emphasis between movement and manipulation with the consideration of the robot's state and environment. In addition, we propose the Monitor-Position-Based Servoing (MPBS) with $\textbf{sig}(\omega)$, enabling the tracking of the target during the operation. This approach enables coordinated control among the robot's base, arm, and camera, enhancing task efficiency. Through extensive simulations and real-world experiments, our approach significantly improves both the success rate and efficiency of mobile manipulation tasks, achieving a 95.6\% success rate in real-world scenarios and a 52.8\% increase in time efficiency.

\end{abstract}

\section{INTRODUCTION}
\begin{figure}[!t]
    \centering
    \begin{subfigure}[b]{0.45\textwidth}
        \includegraphics[width=\textwidth]{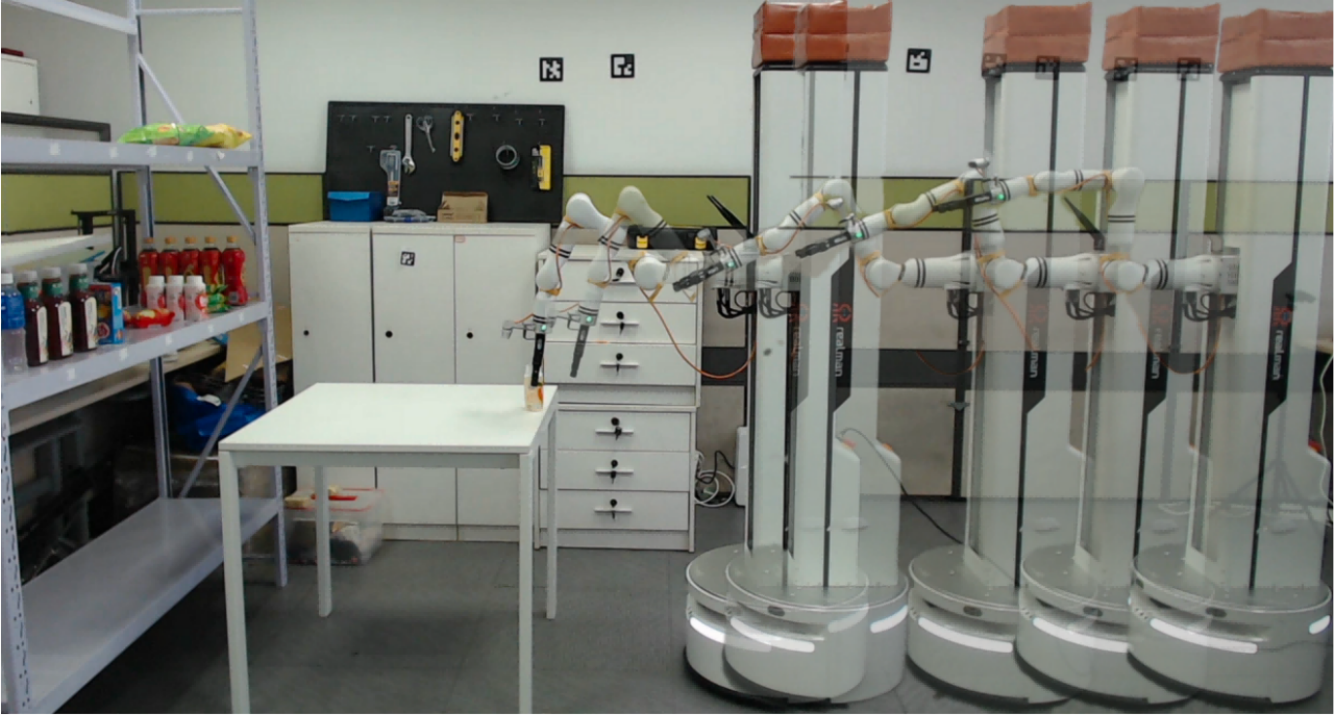}
        \caption{Real-world scenarios}
        \label{teaser:simulation}
    \end{subfigure}
    \hfill 
    \begin{subfigure}[b]{0.45\textwidth}
        \includegraphics[width=\textwidth]{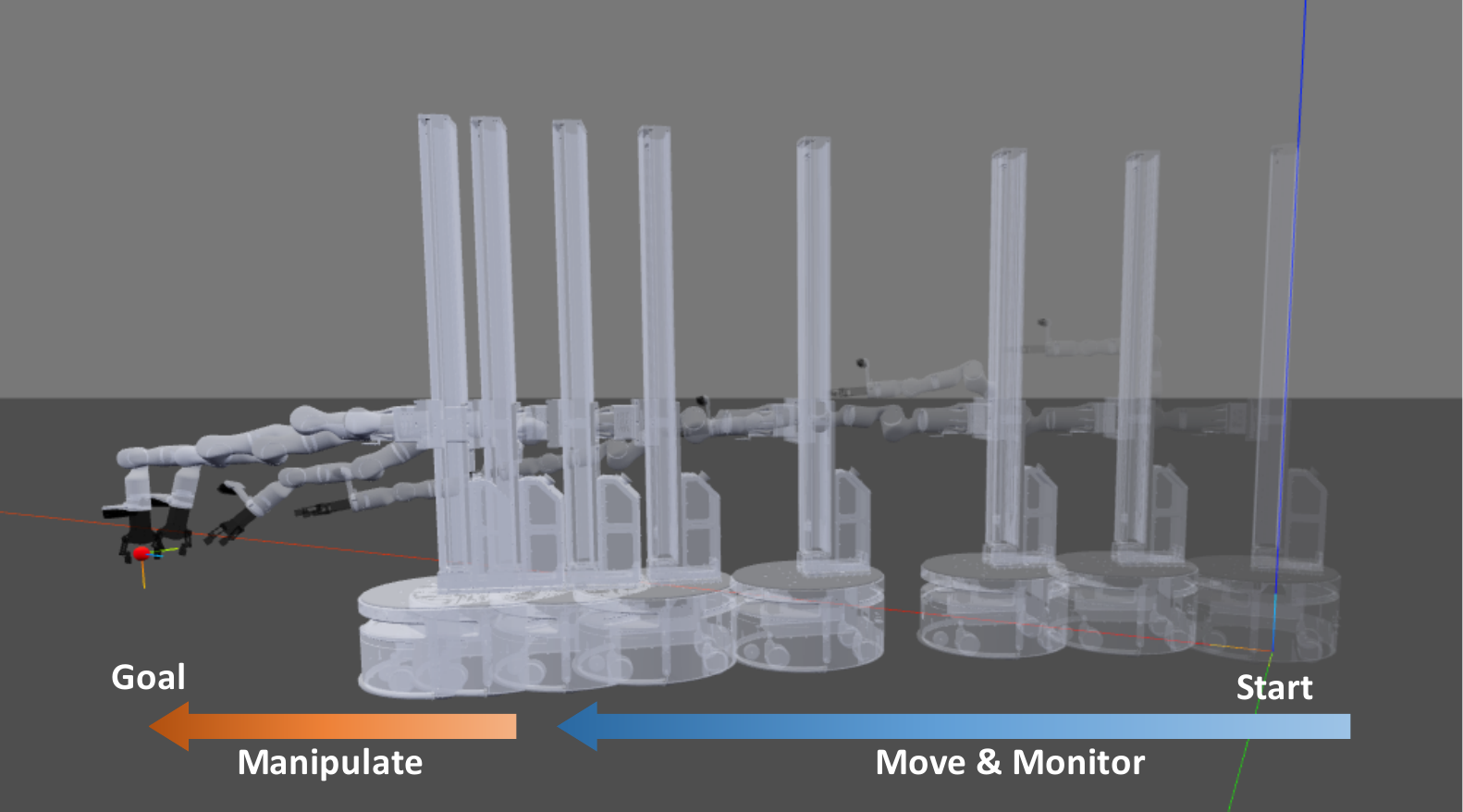}
        \caption{Simulation}
        \label{teaser:real}
    \end{subfigure}
    \caption{\textbf{EHC-MM.} The function $\textbf{sig}(\omega)$ of EHC-MM dynamically balances the robot’s emphasis between movement and manipulation with the consideration of the robot's state and environment. It coordinates all joints of the robot simultaneously and has been validated both in simulation and real-world scenarios.}
    \label{paper-show}
\end{figure}

Mobile manipulation is expected to play a significant role in industrial and domestic applications. However, in many practical robot applications, the planning for the mobile base and robot arm is often separated. This separation in planning processes can lead to an unnatural approach, resulting in the loss of optimal solutions\cite{jiao2021efficient}. Consequently, this approach significantly increases the failure rate of planning and proves unsuitable for complex, unstructured environments. For mobile manipulation, Holistic Control, also known as Whole Body Control is essential and indispensable. Presently, in the context of mobile manipulation, two distinct challenges persist concerning the implementation of Holistic Control in physical robotic motions:

\textbf{Lack of proprioception}. Previous works mostly treat the entire robot as a single entity without considering the functional differences among its joints\cite{jiao2021efficient, Shiqi2023, cho2022omega,jiao2021consolidating}. 
Therefore, it's crucial to coordinate control between the base and arm by considering the mobility and accuracy of each part. The mobile base has greater mobility but lacks accuracy; the robot arm offers higher accuracy in manipulation but has limited workspace. That is why, for tasks involving movement and manipulation, we aim for the robot to the \textit{Distant Mobility, Close Grasping} (DMCG) principle: prioritizing base movement when at a distance and focusing on arm manipulation when close to the target \cite{haviland2022holistic}, as shown in Fig. \ref{paper-show}. Given inevitable perception and motion errors in the real world, integrating closed-loop systems with real-time planning and servoing is essential\cite{haviland2022holistic,Mittal_Hoeller_Farshidian_Hutter_Garg_2022}.

\textbf{Perception-motion coordination}. Traditionally, visual sensing and robotic manipulation are combined in an open-loop fashion by first ‘perceiving’ and then ‘moving’\cite{Ken2017, Ashutosh2015}. For unstructured unknown environments, visual perception relies entirely on cameras mounted on mobile manipulation robots\cite{Akansel,angelopoulos2023high}. However, when camera is mounted at the end of the robot arm like Fig. \ref{paper-show} or on the top of the robot, there exists a deficiency in effective perception and motion coordination mechanisms.

For the existing challenges, we propose the Embodied Holistic Control for Mobile Manipulation(EHC-MM), as shown in Fig. \ref{fig:framework}. 

Our main contributions include:

\begin{itemize}
    \item We formulate the DMCG problem in mobile manipulation as a quadratic programming problem, incorporating both obstacle avoidance and joint angle constraints. This enables efficient simultaneous planning of all robot joints.

    \item We design the $\textbf{sig}(\omega)$ to achieve embodied control. Specifically, $\textbf{sig}(\omega)$ accounts for both the reachability of the robot’s joints and the task objectives, allowing the robot to balance between movement and manipulation. Additionally, we developed a monitor-position-based servoing (MPBS) function to prevent the robot from losing track of the target during its operation.

    \item We conduct comprehensive comparative experiments and ablation studies in both simulation and real-world scenarios. The results validate that our approach achieves higher success rates and efficiency in grasping.
\end{itemize}

\section{RELATED WORK}

\subsection{Mobile Manipulation}
Mobile manipulation is extensively explored across various domains in recent years. Notably, Chitta et al. \cite{Chitta_Jones_Ciocarlie_Hsiao_2012} utilize visual, tactile, and proprioceptive sensor data to execute pick-and-place tasks with the PR2 robot. In addition, HERB \cite{Srinivasa_Ferguson_Helfrich_Berenson_Collet_Diankov_Gallagher_Hollinger_Kuffner_Weghe_2010} enables a robotic agent to identify grasping targets and navigate through complex environments. These contributions lay the foundation for mobile manipulation by emphasizing the segmentation of moving and manipulation tasks. However, this partitioning in the planning domain is inherently artificial and introduces unnecessary complexity into the planning space.

To address the challenge of whole-body control in mobile grasping, Haviland et al. \cite{haviland2022holistic} adapt NEO \cite{neo} from stationary to mobile manipulators, proposing a novel taskable reactive mobile manipulation system. They treat the arm and base degrees of freedom as a unified structure, significantly enhancing the speed and fluidity of motion.

Drawing inspiration from the Virtual Kinematic Chain (VKC) concept \cite{Pratt_Dilworth_Pratt_2002}, Jiao et al. \cite{jiao2021efficient, jiao2021consolidating} develop a Virtual Kinematic Chain framework for mobile manipulation tasks. This approach integrates the kinematics of the mobile base, robot arm, and manipulated object, simplifies task planning by defining abstract actions, and reduces complexity in describing intermediate poses. Implementing a task planner using PDDL with VKC shows significant improvements in planning efficiency, with experiments demonstrating the effectiveness of VKC-based planning in generating feasible motion plans and trajectories.

\subsection{Embodied Holistic Control}
Previous work has introduced the concept of Embodied Control into optimization control of quadrupedal robots\cite{emboiedqu}, swarm robots\cite{embodiedswarm}, soft robots\cite{embodiedzhang}, and human-robot interaction\cite{embodiedhri,embodiedhri1}. This approach has streamlined control system design, enhancing robot adaptability and stability.

Regarding the distinct roles of different robot components in mobile manipulation, each part serves a specific function. Typically, the mobile base exhibits lower precision, on the order of centimeters, but facilitates spatial mobility. In contrast, the robot arm offers higher precision, operating at the millimeter level, albeit within a limited workspace, primarily for object manipulation tasks, as shown in Fig. \ref{paper-show}

\begin{figure*}[!ht]
    \centering
    \includegraphics[width = 0.95 \textwidth]{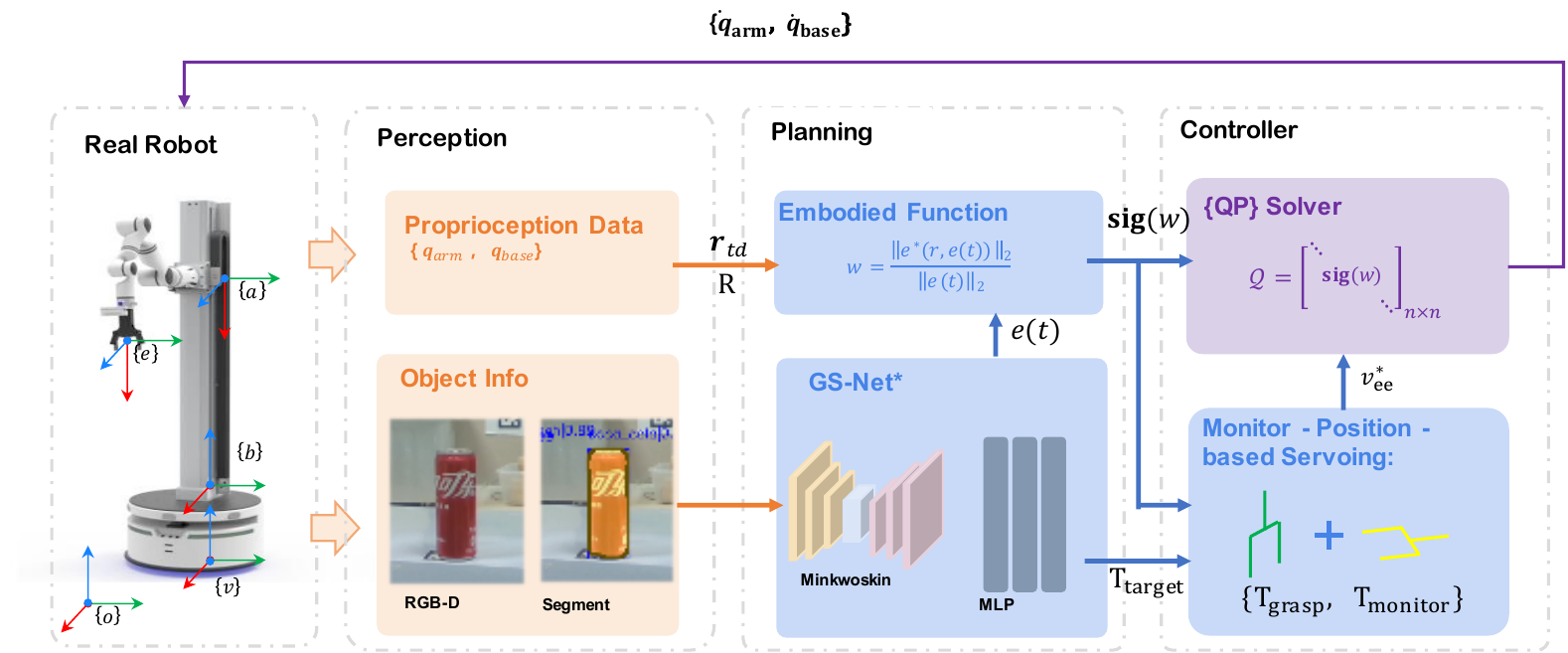}
    \caption{\textbf{Framework of EHC-MM.} The core of the framework is the embodied function $\textbf{sig}(\omega)$, which dynamically balances the robot’s focus between movement and manipulation, taking into account both the robot’s state and its environment. Combined with MPBS, $\textbf{sig}(\omega)$ improves the robot’s ability to track targets. Moreover, our proposed GS-Net* enhances the stability of grasp poses. Overall, this framework significantly improves the efficiency of mobile manipulation.}
    \label{fig:framework}
\end{figure*}

\section{ARCHITECTURE of EHC-MM}

We present an architectural overview in Fig. \ref{fig:framework}. We divide the entire mobile manipulation into four parts: \textit{Real Robot}, \textit{Target Perception}, \textit{Motion Planning} and \textit{Controller} which are commonly observed in most robot frameworks. Our enhancements primarily focus within the blue rectangular box depicted in Fig. \ref{fig:framework}.

\subsection{Kinematic Model and Control of the Mobile Manipulation}

According to the chain rule in robot kinematics, the pose of the robot's end-effector can be represented as:

\begin{equation}
    {}_{e}^{o}\mathbf{T}={}_{v}^{o}\mathbf{T}(x, y,\alpha ) \cdot {}_{b}^{v}\mathbf{T} \cdot {}_{a}^{b}\mathbf{T} \cdot {}_{e}^{a}\mathbf{T} \in \textbf{SE}(3)
\end{equation} 
where \{o\} is the world reference frame. The matrix ${}_{v}^{o}\mathbf{T}$ denotes the transformation of the virtual base frame \{v\} to the world frame, through two perpendicular prismatic joints and a revolute joint. $(x, y, \alpha)$ represents the mobile base's position and orientation in the 2-D plane. ${}_{b}^{v}\mathbf{T}$ represents a constant transformation of the mobile base \{b\} to the virtual base frame on the robot.  ${}_{a}^{b}\mathbf{T}$ signifies a constant relative transformation from the base frame to the base of the manipulator arm \{a\}, and ${}_{e}^{a}\mathbf{T}$ represents the forward kinematics of the arm where the end-effector frame is \{e\}. This configuration is illustrated in Fig. \ref{fig:framework}.

The resulting expression could also be written as the following differential kinematics Equation \ref{ve}, where $\dot{\boldsymbol{\theta}}(t)$ represents the joint velocity vector of the robot.
\begin{align}
{ }^{o} \boldsymbol{\nu}_{e} 
& ={ }^{o} \mathbf{J}_{e}(x, y, \alpha, \boldsymbol{\theta}(t)) \dot{\boldsymbol{\theta}}(t) \label{ve} 
\\
\dot{\boldsymbol{\theta}}(t) &= \left(\begin{array}{c}\dot{\boldsymbol{q}}_{base} \\ \dot{\boldsymbol{q}}_{arm}\end{array}\right) 
\end{align}

For obstacle avoidance, building on previous works\cite{neo, li2024quadratic}, a minimum distance $d$ exists between a point $L$ on the robot's link and a point $O$ on the obstacle in 3D space. The goal is to ensure that $d \ge S$, where $S$ represents a predefined safety threshold. If $d$ falls below $S$, a collision with the obstacle is considered to have occurred. Thus, the collision-free criterion is defined as follows:
\begin{equation}
     \left \| \boldsymbol{p_L}(\boldsymbol{\theta}(t)) - \boldsymbol{p_O}(t) \right \|_2 \ge S
     \label{eqoa}
\end{equation}
where $\boldsymbol{p_L}(\boldsymbol{\theta}(t)) \in \mathbb{R}^{3}$ and $\boldsymbol{p_O}(t) \in \mathbb{R}^{3}$ represent the coordinates of point $L$ and point $O$, respectively. Evidently, inequality \ref{eqoa} is equivalent to
\begin{equation}
      (\boldsymbol{p_L}(\boldsymbol{\theta}(t)) - \boldsymbol{p_O}(t))^{\top} (\boldsymbol{p_L}(\boldsymbol{\theta}(t)) - \boldsymbol{p_O}(t)) \ge S^2
\label{ss}
\end{equation}

Building upon the model introduced by Jesse Haviland et al. \cite{haviland2022holistic, neo}, 
we have further advanced the controller for mobile manipulation tasks by reformulating it as a quadratic programming (QP) problem, as described in Equation \ref{qp}.
\begin{align}
\min _{x} \quad f(\boldsymbol{x}) & =\frac{1}{2} \boldsymbol{x}^{\top} \mathcal{Q} \boldsymbol{x}+\mathcal{C}^{\top} \boldsymbol{x} \label{qp}\\
\text { subject to } \quad \mathcal{J} \boldsymbol{x} \ & = \boldsymbol{\nu}_{e} - \boldsymbol{\delta} (t) \nonumber \\
G(\boldsymbol{x}) & \leq \boldsymbol{r}_I  \nonumber
\end{align}
where $\boldsymbol{x} = \dot{\boldsymbol{\theta}}(t) ^{\top}$ is the decision variable, $\boldsymbol{\delta} (t)$ represents the difference between the desired and actual end-effector velocity, which relaxes the trajectory constraint. $\mathcal{Q}$ incorporates joint velocity between the mobile base and the robot arm, $\mathcal{C}$ incorporates manipulability maximisation and auxiliary performance tasks, $\mathcal{J}$ incorporates an augmented manipulator Jacobian. 

The constraint function $G(\boldsymbol{x})$ incorporates Equation \ref{ss} and joint velocity limits, and its upper bound $\boldsymbol{r}_I$ is defined as:
\begin{align}
    G(\boldsymbol{x}) &= \begin{bmatrix}
-H(\boldsymbol{x})^{\top} H(\boldsymbol{x}) \\
\boldsymbol{x}\\
- \boldsymbol{x}
\end{bmatrix} \in \mathbb{R}^k \\
\boldsymbol{r}_I &= \begin{bmatrix}
-S^2 \\
\mathcal{X}^{+}\\
-\mathcal{X}^{-}
\end{bmatrix}\in \mathbb{R}^k 
\end{align}
where $\mathcal{X}^{+,-}$ limits maximum and minimum values of the decision variable. The function $H(\boldsymbol{x})$ describes the relative position relationship between a point in the robot link and the obstacle, as defined in Equation \ref{Hx}.
\begin{equation}
H(\boldsymbol{x}) = \left(\boldsymbol{p_L}\left(\boldsymbol{\theta}(0) + \int_0^t \boldsymbol{x} , dt\right) - \boldsymbol{p_O^{rel}}(t)\right)
\label{Hx}
\end{equation}
where $\boldsymbol{p_O^{rel}}(t)$ represents the relative position of the obstacle point $O$ to the virtual base, computed as:
\begin{equation}
\boldsymbol{p_O^{rel}}(t) = \beta \psi\left({ }^o_v \mathbf{T}^{-1}\left(x, y, \theta\right) \cdot {}_{obs}^{\ o}\mathbf{T}\right)
\end{equation}
Here, ${}_{obs}^{\ \ o}\mathbf{T} \in \textbf{SE}(3) $ is the obstacle in the virtual base frame. $\beta \in \mathbb{R}^+$ is a gain term, and $\psi (\cdot ):\mathbb{R} ^{4\times 4}  \mapsto  \mathbb{R} ^{6}$ is a function which converts a homogeneous transformation matrix to a spatial displacement.

\subsection{EHC Function }

For mobile manipulation, the accuracy and motion abilities of each joint exhibit variations. Generally, robot arms exhibit higher accuracy but have limited workspace and higher computational complexity. While the mobile base extends the operational workspace significantly, it generally shows a greater degree of motion inaccuracy compared to robot arms. Hence, we designed the Embodied Holistic Control(EHC) Function, as shown below:

\begin{align} 
 \mathcal{Q} 
& =\left(\begin{array}{cc}\operatorname{diag}\left( \textbf{sig}(\omega)\right)  & \mathbf{0}_{n_{base} \times n_{arm}} \\ \mathbf{0}_{n_{arm} \times n_{base}}&\mathbf{I} _{n_{arm}} \\ \end{array}\right) \\
 \mathcal{C} & =\left(\begin{array}{c}\mathbf{J}_{m}\\ \mathbf{0}_{6 \times 1}\end{array}\right)\\
\mathcal{Q} & \in \, \mathbb{R}^{(n+6) \times(n+6)},\
\mathcal{C}  \in \, \mathbb{R}^{(n+6)} \nonumber
\end{align}
where
\begin{align}
\omega \ \ &= \frac{\left \|{e}^{*}(r, e(t))\right \|_2}{\left \| e(t) \right \|_2} \\
 {e}^{*} (r, e(t)) &= \mathop{\arg\min}_{r \in \boldsymbol{r}_{td}} \angle(r, e(t)) \\
 \boldsymbol{r}_{td} \ \ &=\{r\in \textbf{SE}(3) | R(r)>0.95\}
\end{align}
$\textbf{sig}(\omega )$ dynamically balances the emphasis between the movement and manipulation, allowing the robot to switch its focus between efficient movement and precise manipulation as required by target pose. In particular, $e(t) \in \textbf{SE}(3)$ represents the end-effector's pose relative to the target from perception in Fig. \ref{fig:framework}. \( \boldsymbol{r}_{td} \) represents the set of end-effector poses with reachability greater than 95\%, while \({e}^{*} (r, e(t))\) denotes the threshold pose that is most closely aligned with the direction of the current target error \(e(t)\), serving as a threshold for motion focus under the current task objective. In addition, $R(\cdot)$ represents the reachability of the robot arm's end-effector. The activation function $\textbf{sig}(\cdot )$ is incorporated into the robot control function, amplifying the emphasis on the specific motion. This enhancement improves the overall efficiency of motion planning. $\mathbf{J}_m$ is the manipulability Jacobian.

By employing this approach,  the robot can effectively adhere to the \textbf{DMCG principle}: When the robot is positioned at a considerable distance from the target object, the mobile base primarily handles the task of approaching the target, while the robot nears the target and within an embodied threshold $\omega$ accounting for the robot’s reachability $\boldsymbol{r}_{td}$, the robot arm takes over as the primary actor in the manipulation process. In particular, this strategy enables the simultaneous planning of both moving and manipulation tasks, representing a significant advancement over traditional two-stage methods.

\subsection{Monitoring-Position-Based Servoing}

\begin{figure}[h!]
  \centering
  \begin{subfigure}[b]{0.49\textwidth}
      \includegraphics[width=\linewidth]{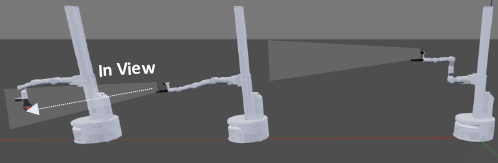}
      \caption{\textbf{Controller upon PBS}. The robot lost the target during the process of reaching.}
      \label{mbs1}
      \end{subfigure}
  \hfill
  \begin{subfigure}[b]{0.49\textwidth}
      \includegraphics[width=\linewidth]{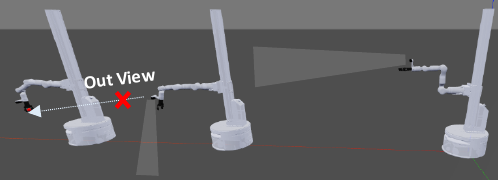}
      \caption{\textbf{Controller upon MPBS}. The robot remains focused on the target throughout the process of reaching.}
      \label{mbs2}
  \end{subfigure}
  \caption{A comparison of the mobile manipulation between PBS and MPBS.}
  \label{fig:mbs}
\end{figure}

As illustrated in the Fig. \ref{mbs1}, current Position-Based Servoing(PBS)\cite{haviland2022holistic} methods face the challenge of target loss during the reaching process, which is particularly detrimental for reactive tasks. This is because, during the process of mobile manipulation, when the grasping orientation is not aligned with the forward-facing direction, such as when the orientation is downward or sideways, the end-effector of the robot arm changes its orientation prematurely during the reaching phase, resulting in a loss of visual contact with the target. Burgess et al.\cite{Burgess-Limerick_Lehnert_Leitner_Corke_2022} propose improvements for grasping objects on the move. However, in their work, the object's movement information is available to the robot. In real-world complex scenarios, the object's positional information is only accessible to the robot through sensors. This necessitates that the robot should maintain the camera's focus on the target during moving.

We utilize a controller that builds upon PBS and incorporates Monitoring-Position-Based Servoing (MPBS). This controller is designed to be parallel with the EHC function that dynamically balances the emphasis between monitoring and manipulation, based on $\textbf{sig}(\omega)$.
The MPBS is formulated as Equation~\ref{mpbs},
\begin{equation}
{ }_{e} ^{b} \mathbf{T}^* = \textbf{sig}(\omega )\cdot { }^b _{e}\mathbf{T}+(1-\textbf{sig}(\omega ))\cdot { }_{m}^{\ b}\mathbf{T}
\label{mpbs}
\end{equation}

where the ${ }_{e} ^{b} \mathbf{T}^ \in \mathbf{SE}(3)$ is the end-effector target pose and ${ }_{m}^{\ b}\mathbf{T} \in \mathbf{SE}(3)$ is the pose for monitoring. $\textbf{sig}(\omega )$ represents the threshold value for switching between the grasping pose and the monitoring pose, as shown in Fig. \ref{paper-show}.

\section{EXPERIMENTS}

First, we validate the effectiveness of our method compared to the baselines in simulation. Then, we compare the continuous grasping performance of our method and the baselines in real-world experiments. Subsequently, we verify the effectiveness of MPBS for improving grasping objects with different poses through ablation.

\subsection{Hardware Setups}
We utilize a RealMan compound robot to conduct experiments. This robot integrates a Water2 non-holonomic mobile base, featuring two drive wheels and four omni-directional wheels, enabling both in-place rotation and straightforward linear motion. The manipulator employed is an RM65-B six-degree-of-freedom robot arm. For visual perception, an Intel Realsense D435 camera is mounted at the end effector of the manipulator. The end effector gripper has a maximum aperture of 70mm and can exert a maximum gripping force of 1.5 kg. The computation is performed on a Alienware laptop, equipped with an Intel(R) Core(TM) i9-9900K CPU and a Nvidia GeForce RTX 2080 GPU.

\subsection{Grasp Pose Generation}
To obtain a high-quality grasp pose, we enhanced GS-Net \cite{wang2021graspness} that only considers the grasping problem of fixed arms in static scenes, and named the improved version GS-Net*.
\begin{figure}[h!t]
    \centering
    \includegraphics[width = 0.45\textwidth]{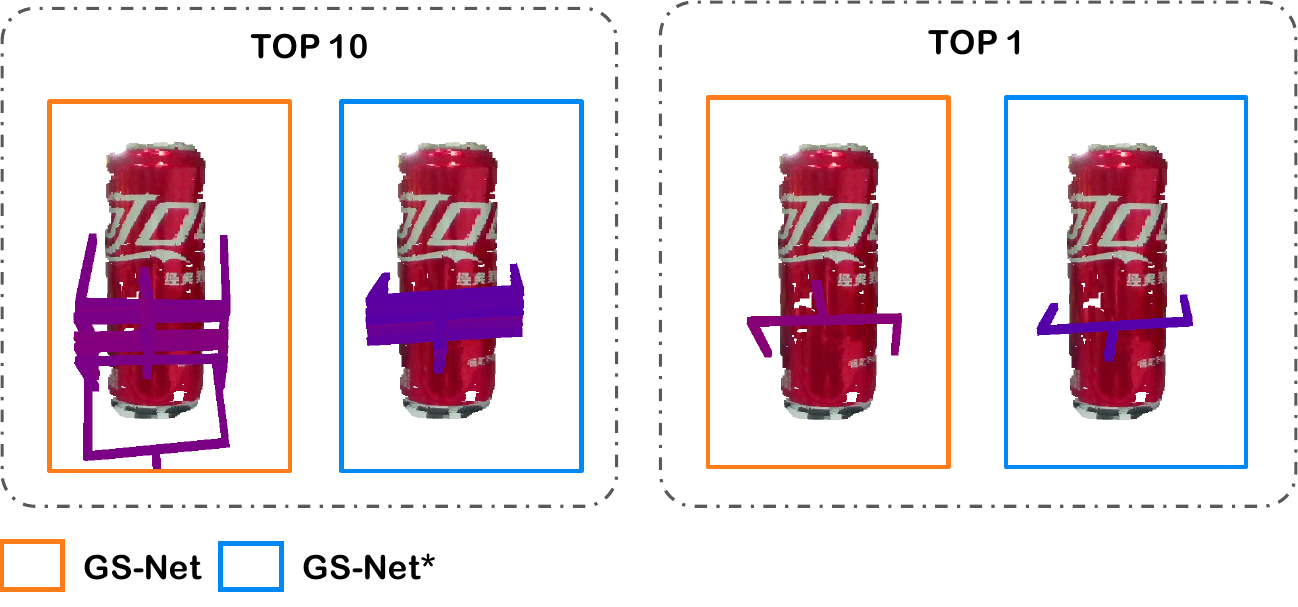}
    \caption{Top-10 and top-1 grasp poses for GS-Net and our GS-Net*}
    \label{gp}
\end{figure}
GS-Net* improves mobile manipulation stability by generating grasp poses aligned with the object's center of gravity and parallel to the object. After the approaching net phase, view scores are adjusted based on the angle to the center of gravity, with smaller angles receiving higher weights. Post-processing aligns the grasp poses with the object's point cloud along the y-axis. Fig. \ref{gp} demonstrates that GS-Net* produces more stable, centered grasp poses. Compared with the original GS-Net, GS-Net* significantly increases the success rate by 28.9\% for the same tasks.

\subsection{Evaluation Metrics and Baselines}
\textbf{Evaluation Metric.} 

\begin{itemize}
    \item {\textit{Time for Grasping}}(\textbf{s}): In simulation, it refers to the average time taken to achieve the target pose. In real-world experiments, it represents the time taken to successfully grasp the target, considered only for successful trials.
    
    \item {\textit{Time for Monitoring}(\textbf{\%}}): The ratio of the time the target object remains within the camera's field of view to the total operation time, referred to as \textbf{\textit{TfM}}.
    
    \item {\textit{Joint Velocity}}(\textbf{rad/s} for rotation, \textbf{m/s} for translation): For the base, which consists of two virtual joints, translation and rotation are considered, denoted as \textbf{Base-T} and \textbf{Base-R}, respectively. For the arm, it represents the combined velocity of the six rotational joints.
    
    \item {\textit{Success Rate}}(\textbf{\%}): In the simulation, it refers to the percentage of trials in which the end-effector successfully reached the target pose. In real-world experiments, it refers to the percentage of trials in which the end-effector successfully grasps the object.
\end{itemize}

\textbf{Baselines.} Our improvements are primarily built on the NEO Controller\cite{haviland2022holistic,neo}, thus, their approach serves as the baseline. The specific configurations are outlined below:
\begin{enumerate}
    \item \textbf{TSMM}: Traditional \textbf{T}wo-\textbf{S}tage \textbf{M}obile \textbf{M}anipulation. The robot first moves to a position \textbf{$d$} away from the target object's position, measured from the end-effector. Then the robot proceeds to grasp the object. In real-world experiments, \textbf{$d$} is set to 0.2.
    \item \textbf{NEO(e)}: The methodology of NEO\cite{haviland2022holistic}, where the weight of the base is simply set to $\tfrac{1}{\left \| e(t) \right \|_2} $.
    \item \textbf{NEO(c)}: To further demonstrate the validity of our work's improvement direction, we adjust the weight of the base in the original method to a constant value $c$. This adjustment serves to validate the correctness of our method's motivation.

\end{enumerate}

\begin{figure}
    \centering
    \setlength{\abovecaptionskip}{5pt} %
    \setlength{\belowcaptionskip}{-15pt}
    \includegraphics[width = 0.5\textwidth]{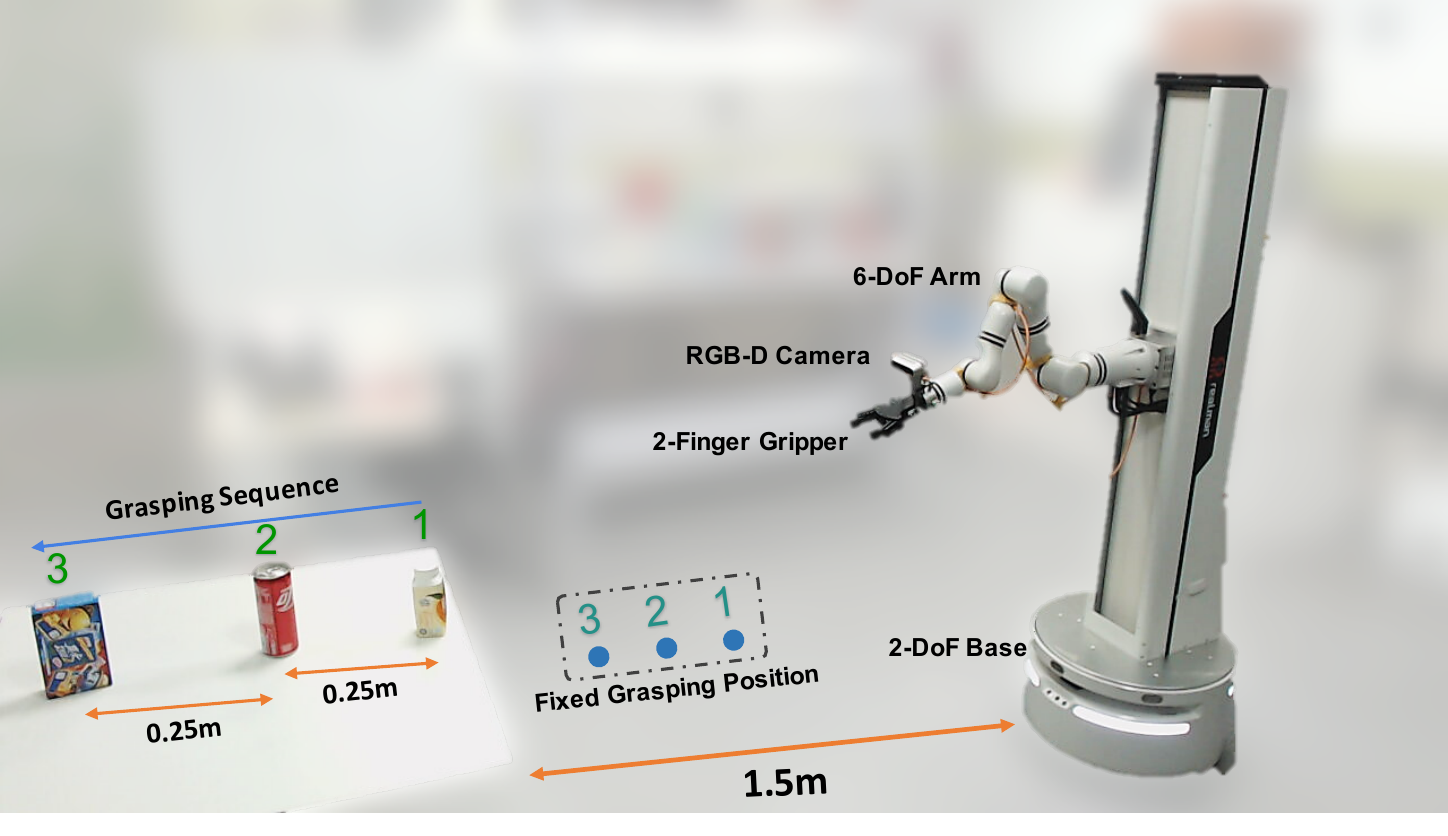}
    \caption{\textbf{Setup of experiment-2.} The Fixed Grasping Position in the figure is only set for baseline \textbf{TSMM}.}
    \label{fig:realworld1}
\end{figure}

\vspace*{-5pt}
\subsection{Experiment-1: Randomly Reaching}

\textbf{Experimental Protocol.}  We conduct this performance test in the simulation. We introduce different levels of noise to various joints, as shown in Table \ref{tab:noise}. In this task, the robot is asked to reach the random 50 points in sequence.

\textbf{Results and Analysis.} We compared \textbf{NEO(c)}, \textbf{NEO(e)}, and \textbf{EHC}. In each set of experiments, the three approaches individually reached the same 50 random points. We conducted a total of 10 sets of experiments, thereby each method approaching 500 random points. The results of the experiment are displayed in the Table \ref{tab:sim}. We find that our approach aligns with the DMCG principle, prioritizing mobile base for moving at a distance, and robot arm for manipulation at close range. Additionally, due to the uncertainty associated with randomly generated points, some target poses may be challenging to reach. Hence, we set a maximum time limit of 30 seconds for each approach attempt. Otherwise, it is considered as a failure. Our method's ability to coordinate movements and manipulation enables the robot to reach the desired target pose in a shorter time, resulting in fewer failures. 

\begin{table}[!htb]
\centering
\caption{The Magnitude of Noise Added to Each Joint.}
{%
\label{tab:noise}
\begin{tabular}{m{1.5cm}<{\centering} m{1cm} <{\centering} m{1cm}<{\centering} m{1cm}}
\toprule
Joints & Base-T & Base-R & Arm  \\ 
\midrule[0.5pt]
Noise & 0.05 & 0.05 & 0.002 \\ \bottomrule
\end{tabular}}
\end{table}

\begin{table*}[!htb]
\centering
\begin{threeparttable}         
\caption{Randomly Reaching in Simulation}
\label{tab:sim}
\begin{tabular}{c c c c c}
\toprule
Method & Time(s) & Distant(Base-T/Base-R/Arm) & Close(Base-T/Base-R/Arm) & \# of Failed (/500)   \\ \midrule[0.5pt]
NEO(c) & 11.69 & 1.644/0.691/5.879 & 0.091/0.062/0.641 & 37 \\ 
\midrule[0pt]
NEO(e) & 11.81 & 1.630/0.683/\textbf{5.870} & 0.087/0.056/0.607 & 35\\ 
\midrule[0pt]
EHC & \textbf{9.98} & \textbf{1.716}/\textbf{0.733}/7.068 & \textbf{0.080}/\textbf{0.042}/\textbf{0.870} & \textbf{18}\\
\bottomrule
\end{tabular}
\footnotesize               %
Notes: bold values indicate the best for the respective metric.
\end{threeparttable}       %
\end{table*}

\begin{table*}[!ht]
\centering
\begin{threeparttable}          %
\setlength{\belowcaptionskip}{-5pt}
\caption{Sequentially Grasping Multiple Objects in Real-World Experiments}

\label{tab:realworld2}
\begin{tabular}{c c c c c}
\toprule
Method & Time(s) & Distant(Base-T/Base-R/Arm) & Close(Base-T/Base-R/Arm) & Success Rate (\%)  \\ \midrule[0.5pt]
TSMM & 34.89 & NAN/NAN/NAN & NAN/NAN/NAN & 80.0 \\ 
\midrule[0pt]
NEO(c) & 30.03 & 8.776/0.578/\textbf{1.971 }& 1.294/0.335/0.345 & 11.1 \\ 
\midrule[0pt]
NEO(e) & 27.80 & 8.709/0.517/2.284& 1.270/0.416/0.638 & 51.1\\ 
\midrule[0pt]
EHC & \textbf{13.54} & \textbf{8.826}/\textbf{0.908}/2.594 & \textbf{1.259}/\textbf{0.162}/\textbf{1.369} & \textbf{95.6}\\ \bottomrule
\end{tabular}
\footnotesize               
Notes: bold values indicate the best for the respective metric.
\end{threeparttable}       
\end{table*}

\begin{table}[!htb]
\centering
\caption{Grasping the Object with Different Poses}
{\label{tab:realworld3}
\begin{tabular}{cccc}
\toprule
\multirow{2}{*}{Method} & \multicolumn{3}{c}{\textit{\textbf{TfM}}(\%)} \\
& \textbf{Forward} & \textbf{Downward}     & \textbf{Sideway}     \\ \midrule[0.5pt]
NEO(without MPBS)  & 100.0     & 31.6     & 38.7     \\
\midrule[0pt]
EHC(without MPBS)& 100.0        & 34.3     & 36.2     \\
\midrule[0pt]
EHC(ours)       & 100.0             & \textbf{100.0}     & \textbf{100.0}      \\ \bottomrule
\end{tabular}}
\end{table}

\vspace*{-5pt}
\subsection{Experiment-2: Sequentially Grasping Multiple Objects}

\textbf{Experimental Protocol.} In the real-world experiment, the robot starts from a fixed location and sequentially grasp three objects of different distances on a table 1.5 metres away. These three objects were placed at the front, middle and back positions on the table, as shown in the Fig. \ref{fig:realworld1}. The time is the average time from the start of the robot's movement, to the completion of the grasp, considering only successful trails.
\begin{figure}[h]
  \centering
  \setlength{\abovecaptionskip}{15pt}
  \setlength{\belowcaptionskip}{-15pt}
    \includegraphics[width = 0.45\textwidth]{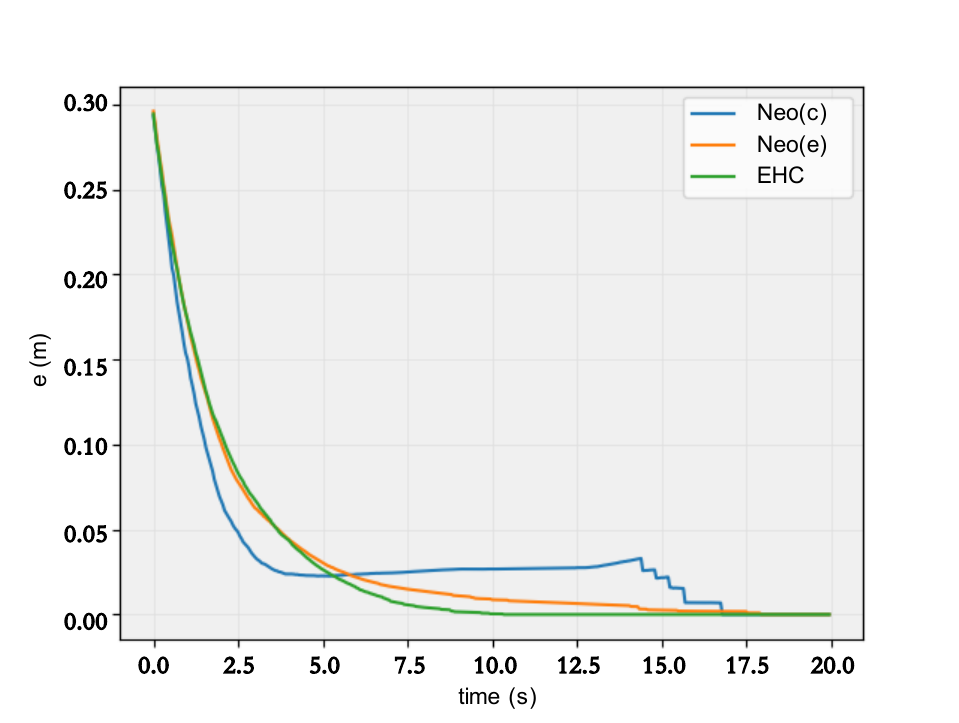}
  \caption{The distance between the end-effector pose and the target pose over time}
  \label{fig:et}
\end{figure}

\textbf{Results and Analysis.} We compare \textbf{TSMM}, \textbf{NEO(c)}, \textbf{NEO(e)} and \textbf{EHC}. We conduct 15 sets of experiments, each grasping 45 objects. The results of the experiment are displayed in Table \ref{tab:realworld2}, where NAN denotes that the value was not measured. Since, for the \textbf{TSMM}, measuring joint changes is meaningless as it includes two separate stages.

Apart from the \textbf{TSMM}, we set a maximum grasping time of 30 seconds for each object. During real-world experiments, we find that the results in the real-world experiments align with those in simulations. \textbf{EHC} achieves shorter grasping times and higher success rates. we observe that the \textbf{NEO(c)} and \textbf{NEO(e)} still frequently perform major movements of the base even close to the target. Due to the limited precision of base movements, the robot faces difficulties in reaching the target smoothly, sometimes leading to failure. As for the \textbf{TSMM}, the robot frequently fails to grasp the most distant object because of the obstruction of the table. From Fig. \ref{fig:et}, we can also visually observe that \textbf{EHC} enables smoother and faster mobile manipulation both in simulation and real-world experiments. Thus, \textbf{EHC} demonstrates excellent performance in real-world scenarios.

\subsection{Experiment-3: Grasping the Object with Different Poses}

\textbf{Experimental Protocol.} In real-world experiment, the robot starts from a fixed location and grasps a single object positioned 1.5 meters away using different grasp poses, including \textbf{forward}, \textbf{downward}, and \textbf{sideways } orientations. The time is the average time cost for each grasping pose.

\textbf{Results and Analysis.} We compared \textbf{NEO(without MPBS)}, \textbf{EHC(without MPBS)} and \textbf{EHC(ours)} as an ablation experiment. 
We conduct 15 sets of experiments, each involving 5 instances of forward, downward, and sideways grasping. The results of the experiment are displayed in the Table \ref{tab:realworld3}. A higher \textbf{\textit{TfM}} value indicates that the robot is less likely to lose track of the target. We observe that, as to \textbf{downward} and \textbf{sideways } grasping orientations, without MPBS, the robot is more likely to lose track of the target in advance, as shown in Fig. \ref{mbs1}. That is because the track of the target is not considered carefully as a constraint in the QP problem. In \textbf{EHC(ours)}, MPBS ensures that the robot maintains monitoring on the target when the robot is at a distance.

\section{CONCLUSIONS}
In this paper, we propose EHC-MM, leveraging the function of $\textbf{sig}(\omega)$. By formulating the DMCG principle as a quadratic programming (QP) problem, we enable simultaneous planning of the robot's joints, rather than relying on traditional two-stage mobile manipulation. The function $\textbf{sig}(\omega)$ dynamically balances the robot's emphasis between movement and manipulation with the consideration of the robot's state and environment, thereby improving both the success rate and efficiency of manipulation tasks. Additionally, $\textbf{sig}(\omega)$ is integrated into the MPBS, allowing the robot to maintain a balance between monitoring and manipulation, ensuring it does not lose track of the target during operation. Through simulation and real-world validation, we observe improved efficiency in mobile manipulation tasks using EHC. In real-world experiments, it achieves an impressive grasp success rate of 95.6\%. The results demonstrate that the proposed method is highly effective for real-world deployments.

\section{ACKNOWLEDGMENT}
The work is supported by the National Natural Science Foundation of China (No. 62176004), Beijing Natural Science Foundation(No. L222012), Wuhan East Lake High-Tech Development Zone(also known as the Optics Valley of China, or OVC) National Comprehensive Experimental Base for Governance of Intelligent Society, Intelligent Robotics and Autonomous Vehicle Lab (RAV), Beijing Samsung Telecom R\&D Center.







\newpage
\bibliographystyle{ieeetr} 
\balance
\bibliography{refs} %

\end{document}